\documentclass[a4paper,oneside,12pt,english]{article}
\usepackage{url}
\usepackage[pdftex]{color,graphicx}
\usepackage[backref=page]{hyperref}
\hypersetup{plainpages = false,
              breaklinks=true,
              linktocpage,
              colorlinks   = true, 
              urlcolor     = blue, 
              linkcolor    = blue, 
              citecolor   = red, 
              pdfauthor={Niko Brummer},
              pdfstartpage=1, 
              pdfstartview=FitH,  
              pdfkeywords={}}

\usepackage{amsmath}%
\usepackage{amsfonts}%
\usepackage{amssymb}%
\usepackage{graphicx}

\usepackage{tikz}
\usetikzlibrary{bayesnet}

\def\Phimat{\boldsymbol{\Phi}}
\def\gammavec{\boldsymbol{\gamma}}
\def\Gammamat{\boldsymbol{\Gamma}}
\def\phivec{\boldsymbol{\phi}}
\def\muvec{\boldsymbol{\mu}}
\def\nulvec{\boldsymbol{0}}

\def\Cmat{\mathbf{C}}
\def\Tmat{\mathbf{T}}
\def\Umat{\mathbf{U}}
\def\xvec{\mathbf{x}}

\def\nulvec{\boldsymbol{0}}
\def\Imat{\mathbf{I}}

\def\ND{\mathcal{N}}
\def\R{\mathbf{R}}

\def\LB{\mathcal{L}}

\DeclareMathOperator{\gmm}{gmm}

\DeclareMathOperator{\softmax}{softmax}

\def\expv#1#2{\left\langle#1\right\rangle_{#2}}

\begin{document}
\title{VB calibration to improve the interface between phone recognizer and i-vector extractor}
\author{Niko Br\"ummer, AGNITIO Research South Africa}
\maketitle

\section{Introduction}
The EM training algorithm of the \emph{classical i-vector extractor}~\cite{ivec1,ivec2} is often incorrectly described as a maximum-likelihood method. The i-vector model is however intractable---the likelihood itself and the hidden-variable posteriors needed for the EM algorithm cannot be computed in closed form. We show here that the classical i-vector extractor recipe is actually a \emph{mean-field variational Bayes (VB)} solution.

This theoretical VB interpretation turns out to be of further use, because it also offers an interpretation of the newer \emph{phonetic i-vector extractor}~\cite{phonetic1, phonetic2}, thereby unifying the two flavours of extractor. 

More importantly, the VB interpretation is also practically useful---it suggests ways of modifying existing i-vector extractors to make them more accurate. In particular, in existing methods, the approximate VB posterior for the GMM states is fixed, while only the parameters of the generative model are adapted. Here we explore the possibility of also mildly adjusting (calibrating) those posteriors, so that they better fit the generative model.

In what follows, we introduce notation by summarizing the i-vector model. Then we interpret the classical recipe to deal with this intractable model as a mean-field VB algorithm. Next, we do the same with the newer phonetic i-vector recipe. Finally, we extend the phonetic recipe by allowing calibration of the phone posteriors. 

\section{The i-vector model}
In what follows, speech segments, indexed by $s$, are represented as variable-length sequences of feature vectors, denoted $\Phimat_s=\{\phivec_{st}\}_{t=1}^{T_s}$, where $\phivec_{st}\in\R^D$ is a feature vector and $T_s$ is the sequence length. The i-vector model is a generative model for these sequences. 

According to the i-vector model, every speech segment is generated by a \emph{different} $N$-component GMM. The GMM for segment $\Phimat_s$ is parametrized by a set of $N$ weights, covariances and means, in the form:
\begin{align}
\gmm(\xvec_s) &= \left\{w_i,\Cmat_i,\muvec_i + \Tmat_i \xvec_s\right\}_{i=1}^N
\end{align}
The weights, $w_i$, and covariances, $\Cmat_i$, remain fixed, while the means vary in a subspace formed by the factor loading matrices, $\Tmat_i$. The result is that every GMM is parametrized by the $M$-dimensional \emph{i-vector}, $\xvec_s$. The special GMM, $\gmm(\xvec=\nulvec)$, has means $\muvec_i$ and is referred to as the \emph{UBM}. 

\begin{figure}[!htb]
\centerline{
\begin{tikzpicture}
\node[latent] (ivec) {$\xvec_s$};
\node[const, outer sep = 15pt] at(ivec|-ivec) (dummy1) {};
\node[obs, right=of ivec] (data) {$\phivec_{st}$};
\node[const, outer sep = 15pt] at(data|-data) (dummy2) {};
\node[latent, right=of data] (state) {$\gammavec_{st}$};
\node[const, outer sep=4pt, above=40pt of data] (T) {$(\Cmat_i,\muvec_i,\Tmat_i)$};
\node[const, outer sep=4pt] at(T-|state) (U) {$w_i$};
\node[const, outer sep=4pt, left=of ivec] (prior) {$\ND(\nulvec,\Imat)$};
\plate[draw=red, inner sep = 5pt] {params} {(U)(T)} {$i$};
\plate[draw=blue,inner sep = 5pt] {innerplate} {(data)(state)(dummy2)} {$t$};
\plate[draw=blue,inner sep = 5pt] {} {(innerplate)(ivec)(dummy1)} {$s$};
\edge {ivec}{data};
\edge {prior}{ivec};
\edge {state}{data};
\edge {T}{data};
\edge {U}{state};
\end{tikzpicture}
}
\caption{\textbf{The i-vector model.} The hidden i-vector is $\xvec_s$, the hidden GMM state is $\gammavec_{st}$ and the observed data point is $\phivec_{st}$. The model parameters are in the red plate.}
\label{fig:model}
\end{figure}
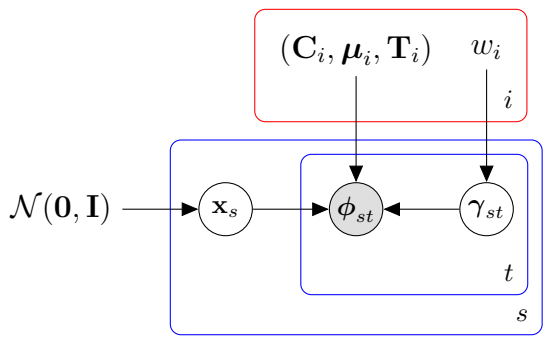

The i-vector model is completed by specifying a standard normal prior on the i-vectors. The complete i-vector model is parametrized by:
\begin{align}
\begin{split}
\Lambda &= \{w_i,\Cmat_i,\muvec_i,\Tmat_i\}_{i=1}^N \\
&= (\Umat,\Tmat)
\end{split}
\end{align} 
where we have grouped the UBM parameters as $\Umat=\{w_i,\Cmat_i,\muvec_i\}_{i=1}^N$ and the factor loading matrices as $\Tmat=\{\Tmat_i\}_{i=1}^N$. The i-vector model is shown in graphical model notation in figure~\ref{fig:model}.

The marginal likelihood, $P(\Phimat_s|\Lambda)$ is intractable, because when only $\Phimat_s$ is given, we cannot jointly sum over the GMM states and integrate out the unknown i-vector. (If the i-vector were also given, the hidden GMM states would factor over frames and could be summed out in closed form. Or if the state path were given, the i-vector could be integrated out in closed form.)

\section{From mean-field VB to the classical i-vector extractor}
\def\const{\mathtt{const}}
For the observed sequence $\Phimat_s=\{\phivec_{st}\}_{t=1}^{T_s}$, let the hidden GMM state for frame $t$ be represented by a one-hot vector $\gammavec_{st}$ and let the whole \emph{path} be denoted $\Gammamat_s=\{\gammavec_{st}\}_{t=1}^{T_s}$. The hidden \emph{i-vector} is $\xvec_s\in\R^M$. 

In the true joint hidden variable posterior, $P(\xvec,\Gammamat|\Phimat_s,\Lambda)$, the i-vector and path are \emph{not} independent, because of explaining away. We obtain the \emph{mean-field VB} approximation~\cite{bishop} by nevertheless forcing such independence in the approximate posterior:
\begin{align}
P(\xvec,\Gammamat\mid\Phimat_s,\Lambda) &\approx Q_s(\Gammamat,\xvec) = Q_s(\Gammamat)Q_s(\xvec)
\end{align}
where the factors $Q_s(\Gammamat)$ and $Q_s(\xvec)$ are properly normalized distributions that can be found via a variational optimization procedure. More specifically, these factors are found by variational maximization of a functional, known as the \emph{VB lower bound}:
\begin{align}
\label{eq:VBL}
\LB_s &= \expv{\log\frac{P(\Phimat_s,\xvec,\Gammamat\mid\Lambda)}{Q_s(\Gammamat)Q_s(\xvec)}}{Q_s(\Gammamat)Q_s(\xvec)}
\end{align}
where our notation indicates expectation w.r.t. the variational posterior. For such a mean-field factorization, the optimal factors are known\footnote{See e.g.\ \cite{bishop}, chapter 10.} to satisfy:
\begin{align}
\label{eq:mfpathQ}
\log Q_s(\Gammamat) &= \int Q_s(\xvec) \log P(\Phimat_s,\Gammamat,\xvec\mid\Lambda)  \, d\xvec + \const\\
\label{eq:mfivecQ}
\log Q_s(\xvec) &= \sum_{\Gammamat} Q_s(\Gammamat) \log P(\Phimat_s,\Gammamat,\xvec\mid\Lambda)  + \const
\end{align}
where $\const$ is a place-holder for any extra terms not dependent on the variable of interest (respectively $\Gammamat$ and $\xvec$). Since these forms are interdependent, this does not solve the optimal factors in closed form---for that you have to iteratively apply the above equations, effectively doing a form of `co-ordinate ascent'. 

Fortunately, by examining these equations in more detail, we can find the functional forms of the optimal factors. We start by expanding~\eqref{eq:mfpathQ} and absorbing the expected value of $\log P(\xvec)$ into $\const$, giving: 
\begin{align}
\begin{split}
\log Q_s(\Gammamat) &= \int Q_s(\xvec) \log P(\Phimat_s,\Gammamat\mid\xvec,\Lambda)\, d\xvec + \const\\
&= \int Q_s(\xvec) \sum_{t=1}^{T_s} \log P(\phivec_{st},\gammavec_{st},\mid\xvec,\Lambda)\, d\xvec + \const\\
&= \sum_{t=1}^T \int Q_s(\xvec) \log P(\phivec_{st},\gammavec_{st},\mid\xvec,\Lambda)\, d\xvec + \const
\end{split}
\end{align}
This is enough detail\footnote{Later in this document, we shall expand this distribution in more detail.} to show that $Q_s(\Gammamat)$ factorizes over $t$, irrespective of the form of $Q_s(\xvec)$. Note that we did not enforce this factorization over $t$ to start with---the original factorization, $Q_s(\Gammamat)Q_s(\xvec)$ is sufficient to induce this further factorization.

Indexing the components of the one-hot vectors, $\gammavec_{st}$, by the superscripts $i$, we can expand this distribution as:
\begin{align}
\label{eq:mfpathQ2}
\log Q_s(\Gammamat_s)  
&= \sum_{t=1}^T \log Q_{st}(\gammavec_{st}) 
= \sum_{t=1}^T \sum_{i=1}^N \gamma_{st}^i\log q_{st}^i
\end{align}
where we have introduced the variational parameters, $q_{st}^i$, which are probabilities that sum to unity over $i$. Following Bishop, we refer to the $q_{st}^i$ as \emph{responsibilities}~\cite{bishop}. 

\subsection{The VB lower bound}
We now expand the lower bound to the form that we shall use for everything that follows. Employing~\eqref{eq:mfpathQ2} in~\eqref{eq:VBL}, the \emph{VB lower bound} expands to:
\begin{align}
\label{eq:VBLE}
\LB_s &= \expv{\log\frac{\ND(\xvec\mid\nulvec,\Imat)}{Q_s(\xvec)}
+\sum_{t=1}^{T_s} \sum_{i=1}^N q_{st}^i \log\frac
{w_i\ND(\phivec_{st}\mid\muvec_i+\Tmat_i\xvec,\Cmat_i)}
{q_{st}^i} }{Q_s(\xvec)} 
\end{align}
This form can be used:
\begin{itemize}
	\item to derive the formula for $Q_s(\xvec)$;. 
	\item to derive the training algorithm for the extractor parameters, $\Tmat_i$, and also for the UBM parameters if we should need to update them;
	\item as the objective function for finding the parameters involved in re-calibrating the responsibilities.
\end{itemize}
Finally, we shall also make use of a \emph{simplified lower bound}, obtained from~\eqref{eq:VBLE} by setting the $\Tmat_i=\nulvec$, which makes $\xvec$ and $Q_s(\xvec)$ irrelevant. This effectively simplifies the i-vector model to a plain GMM and the resulting lower bound is the same as the well-known EM lower bound for training a GMM:
\begin{align}
\label{eq:LBGMM}
\LB_s^{\nulvec}&= 
\sum_{t=1}^{T_s} \sum_{i=1}^N q_{st}^i \log\frac
{w_i\ND(\phivec_{st}\mid\muvec_i,\Cmat_i)}
{q_{st}^i} 
\end{align}

\subsection{The i-vector posterior}
Given the responsibilities, $q_{st}^i$, we can use~\eqref{eq:mfpathQ2} in~\eqref{eq:mfivecQ} to derive the optimal approximate i-vector posterior:
\begin{align}
\label{eq:mfivecQ2}
\begin{split}
\log Q_s(\xvec) &= \sum_{\Gammamat} Q_s(\Gammamat) \log P(\Phimat_s,\Gammamat,\xvec\mid\Lambda)  + \const \\
&= \log \ND(\xvec\mid\nulvec,\Imat) + \sum_{t=1}^{T_s} \sum_{i=1}^N q_{st}^i \log \ND(\phivec_{st}\mid \muvec_i+\Tmat_i\xvec,\Cmat_i) + \const \\
&= \xvec'\left[\sum_{t,i} q_{st}^i \Tmat'_i\Cmat_i^{-1}(\phivec_{st}-\muvec_i)\right]
-\frac12\xvec'\left[\Imat+\sum_{t,i} q_{st}^i \Tmat'_i\Cmat_i^{-1}\Tmat_i\right]\xvec
+\const
\end{split}
\end{align}
This is a multivariate Gaussian, identical in form to the well-known classical i-vector posterior~\cite{ivec2}. The factors in square brackets are the natural parameters of the Gaussian: the \emph{natural mean} (precision times mean); and the \emph{precision} (inverse covariance).

Although we have derived $Q_s(\xvec)$ from~\eqref{eq:mfivecQ}, the same result could be obtained by working directly from the lower bound~\eqref{eq:VBLE}. To do this, \eqref{eq:VBLE}~can be interpreted (modulo some constants irrelevant to the optimization) as a negative KL divergence from $Q_s(\xvec)$ to the solution~\eqref{eq:mfivecQ2}. Adopting that solution minimizes the KL divergence and thereby maximizes the lower bound. 

\subsection{The i-vector extractor}
As shown above, given the model parameters, $\Lambda=(\Umat,\Tmat)$, the \emph{full} mean-field VB recipe for inferring the i-vector would iteratively update the $q_{st}^i$ and $Q_s(\xvec)$, using~\eqref{eq:mfpathQ} and~\eqref{eq:mfivecQ2}, until convergence. 

The \emph{classical i-vector extractor algorithm} approximates the full recipe with a non-iterative shortcut: 
\begin{enumerate}
	\item \emph{Fix} the responsibilities by equating them to the UBM state posteriors. This is the same as optimizing $\LB_s^{\nulvec}$ w.r.t.\ the $q_{st}^i$ and can be done in one closed-form `E-step'. This step is usually called the \emph{UBM alignment}.
	\item Compute $Q_s(\xvec)$, using~\eqref{eq:mfivecQ2}.
\end{enumerate}

\subsection{The extractor training algorithm}
To train the model parameters, $\Lambda$, using the \emph{full mean-field VBEM}, would involve iterative application of three steps:\footnote{not necessarily in that order}
\begin{itemize}
	\item Update the $q_{st}^i$, using~\eqref{eq:mfpathQ}.
	\item Update the $Q_s(\xvec)$, using~\eqref{eq:mfivecQ2}.
	\item Update $\Lambda$, as the value that maximizes $\sum_s \LB_s$. 
\end{itemize}
In this mean-field VBEM algorithm, the lower bound and all the updates can be computed in closed form.

The \emph{classical `EM-algorithm'} for training the extractor parameters simplifies the full VBEM algorithm by an approximating short-cut: 
\begin{enumerate}
	\item \emph{Fix} $\Umat$ by training an ordinary GMM on the training data, without considering any i-vector mechanism.
	\item \emph{Fix} the $q_{st}^i$, using the same UBM alignment as above. 
	\item After some appropriate initialization, iterate: 
	\begin{itemize}
		\item Update the $Q_s(\xvec)$, using~\eqref{eq:mfivecQ2} 
		\item Update $\Tmat$, by maximizing $\sum_s \LB_s$.  
	\end{itemize}
\end{enumerate}

\section{The phonetic i-vector extractor as a VB algorithm}
In the classical recipe above, we (i) fixed the UBM, which then (ii) supplied the responsibilities, which allowed (iii) training $\Tmat$. Here, we swap the order of steps (i) and (ii), by first fixing the responsibilities via some other means, then finding the UBM parameters and then $\Tmat$.

The fact that the data is observed, places strong constraints on probability distributions for the data. Things like the dimensionality of the data are fixed by the observation. With the hidden variables, we have much more freedom. A wide range of different shapes and sizes of hidden variables can all provide reasonably good explanations for the observed data. Within limits, we could explore different hidden-variable dimensionalities, with different kinds of support (e.g. discrete vs continuous). 

In the case of a the \emph{phonetic i-vector extractor}, we keep the state discrete, but we usually increase $N$ somewhat. We also force the hidden states to correspond to triphone classes (senones). In short, we run a senone classifier on the \emph{raw} speech segment (not on $\Phi_s$, the phone recognizer computes its own features) and the result is a factorial posterior distribution\footnote{With wide input contexts and phonetic and language models, the senone posteriors should not be factorial, but current technology does produce factorial senone posteriors, relying on later processing to link them.} over senone classes for every frame, with the frame rate corresponding to that of the original i-vector extractor. 

In the case of text-independent speaker (or language) recognition, the phone recognizer could be as complex as a full LVCSR system, including a language model; and for text-dependent speaker recognition, it could use forced alignment, using the known text. Or, it could simply be a raw phone recognizer. These details are not important for this analysis, as long as the resulting phone (senone) posterior is factorial over frames. 

The phonetic i-vector extractor recipe can be summarized as follows:
\begin{enumerate}
	\item \emph{Fix} the responsibilities, $q_{st}^i$, by running a senone recognizer on the speech.
	\item \emph{Fix} the UBM parameters, $\Umat=\{w_i,\muvec_i,\Cmat_i\}_{i=1}^N$ by maximizing $\sum_s\LB_s^{\nulvec}$. This can be done in closed form in a single `M-step'.\footnote{Although this also updates the $w_i$, we don't need them for this recipe. We shall however need them for the next recipe when we recalibrate the responsibilities.} 
	\item As above: initialize, then iterate: 
	\begin{itemize}
		\item Update the $Q_s(\xvec)$, using~\eqref{eq:mfivecQ2} 
		\item Update $\Tmat$, by maximizing the full objective, $\sum_s \LB_s$.  
	\end{itemize}
\end{enumerate}

In the above recipe, $\Umat$ is fixed. With a bigger computational budget, it is however possible to maximize $\sum_s \LB_s$ jointly w.r.t.\ $\Umat$ and $\Tmat$ with an iterative EM-algorithm, with closed-form updates. In this EM-algorithm, the E-step updates the i-vector posterior in closed form, while the M-step updates both $\Tmat$ and $\Umat$ in closed form. This may be an interesting experiment to perform---we could potentially obtain a more accurate model, with the caveat that the risk of overfitting increases.

\section{Recalibrating the phone posteriors with VB}
Now we employ the VB interpretation to generalize the phonetic i-vector extractor training, with the purpose of improving the `collaboration' between the phone recognizer and the rest of the extractor. In both of the above VB algorithms (classical and phonetic), the approximate posterior was fixed before finalizing all of the extractor parameters. However, the general VB framework allows for joint optimization of approximate posterior and model parameters. 

In the case of the phone recognizer, we don't want to make big changes to the phone recognizer (presumably it is already working well, and it might be a third party product which is difficult to adjust). But, we can allow calibration of the phone posteriors. This is relatively easy to do and preserves the information contained in the original posteriors. By optimizing the calibration via the VB lower bound, the KL divergence between approximate and actual posteriors will be reduced, allowing them to work better together.

\subsection{The calibration transformation}
In this case, we denote the \emph{fixed} posteriors supplied by the phone recognizer as, $\tilde q_{st}^{\,i}$. Now we allow our VB responsibilities, $q_{st}^i$, to be \emph{adjustable}, but only mildly so. That is, we constrain the responsibilities by deriving them from the fixed phone posteriors via the \emph{calibration transformation}:
\begin{align}
\label{eq:caltrans}
\{q_{st}^i\}_{i=1}^N &= \softmax\left(\{\alpha \log \tilde q_{st}^{\,i} + \beta_i\}_{i=1}^N\right)
\end{align}  
where only the \emph{calibration parameters}, $C=(\alpha,\{\beta_i\}_{i=1}^N)$ are adjustable. The softmax function renormalizes the distribution after the adjustment in the log domain. In the VB context, $C$ may be referred to as a \emph{variational} parameter. The scale factor $\alpha$ adjusts the sharpness of the distribution, while the offsets $\beta_i$ play the same role as adjusting the prior. 

\subsection{The calibration objective}
Here we want to update the responsibilities, $q_{st}^i$, by adjusting the calibration parameters, when the model parameters, $\Lambda=(\Umat,\Tmat)$, and the $Q_s(\xvec)$ are given. To do this, we conveniently massage the lower bound~\eqref{eq:VBLE} into the form of a KL divergence.  It is understood that the maximization below is w.r.t. the $q_{st}^i$ only and that the maximand may be conveniently adapted from line to line by adding or subtracting terms independent of $q_{st}^i$:
\begin{align}
\label{eq:CO}
\begin{split}
&\;\;\;\;\max \sum_s \LB_s \\
&= \max \sum_s 
\sum_{t=1}^{T_s} \sum_{i=1}^N q_{st}^i \expv{\log\frac
{w_i\ND(\phivec_{st}\mid\muvec_i+\Tmat_i\xvec,\Cmat_i)}
{q_{st}^i} }{Q_s(\xvec)} \\
&= \max \sum_s 
\sum_{t=1}^{T_s} \sum_{i=1}^N q_{st}^i \log\frac
{w_i\ell_{st}^i}
{q_{st}^i}  \\
&= \max \sum_s 
\sum_{t=1}^{T_s} \sum_{i=1}^N q_{st}^i \log\frac
{r_{st}^i}
{q_{st}^i}  \\
\end{split}
\end{align}
where we have introduced the likelihoods,
\begin{align}
\label{eq:lli}
\ell_{st}^i &= w_i\exp\left[\expv{\log\ND(\phivec_{st}\mid\muvec_i+\Tmat_i\xvec,\Cmat_i)}{Q_s(\xvec)}\right]
\end{align}
which when normalized give the \emph{optimal responsibilities}:
\begin{align}
r_{st}^i &= \frac{\ell_{st}^i}{\sum_{j=1}^N \ell_{st}^j}
\end{align}
By this normalization, the objective is now in the form of a sum of negative KL divergences, each of which can be minimized (zeroed) by setting $q_{st}^i=r_{st}^i$. If we were doing the full, unconstrained mean-field VB iteration, this is exactly what we would do. Here however, we want to retain the correspondence between senones and our hidden states, so we place a strong constraint, namely~\eqref{eq:caltrans}, on the responsibilities.

In the practical implementation of this calibration it is good to see that~\eqref{eq:lli} can be computed in closed form, since it is a Gaussian expectation of a quadratic expression. To optimize~\eqref{eq:CO} w.r.t.\ the calibration parameters $C$, we will have to resort to general purpose, iterative, numerical optimization, for example BFGS~\cite{nocedal}. 

\subsection{The algoritm}
We can now assemble the full algorithm:
\begin{enumerate}
	\item \emph{Fix} the senone posteriors $\tilde q_{ts}^{\,i}$.
	\item Initialize $C$: $\alpha=1$ and $\beta_i=0$. This implies also, $q_{st}^i=\tilde q_{st}^{\,i}$.
	\item Initialize $\Umat$ as before, by maximizing $\sum_s \LB_s^{\nulvec}$.
	\item Initialize $\Tmat$ and iterate:
	\begin{itemize}
		\item Update $Q_s(\xvec)$, using~\eqref{eq:mfivecQ2}.
		\item Update $C$ using iterative numerical maximization of~\eqref{eq:CO}. 
		\item Update $\Tmat$ and $\Umat$ by maximizing $\sum_s \LB_s$.
	\end{itemize}
\end{enumerate}

\section{Beyond calibration}
As noted above, with hidden variables, we have much freedom. We could generalize the calibration recipe above in various ways:
\begin{itemize}
	\item Generalize $\alpha$ by a diagonal matrix, or a full square matrix, or even a rectangular matrix. (A rectangular matrix can reduce the number of senone classes to a smaller number of hidden variable states.)
	\item Fuse more than one phone recognizer (e.g.\ different ones for different languages), by weighting the log posteriors of each with its own rectangular matrix. With rectangular matrices, we can fuse log posteriors of different sizes.
	\item Smoothing log-posteriors over time, by `fusing' $q_{ti}$ with posteriors of neighbouring frames.
	\item We could also go in the other direction and use VB to design light-weight hidden variable posteriors calculators. One way to do this would be to first find the best-performing (presumably computationally expensive) posterior calculator, with its associated model parameters, $\Umat$ and $\Tmat$. Then fix these model parameters and optimize the (variational) parameters of a computationally cheaper posterior calculator. VB will ensure that the cheaper calculator matches the model parameters as well as possible.
\end{itemize}

\bibliographystyle{IEEEtran}

\end{document}